\pdfoutput=1

\documentclass[11pt]{article}

\usepackage{acl}

\usepackage{times}
\usepackage{latexsym}

\usepackage[T1]{fontenc}

\usepackage[utf8]{inputenc}

\usepackage{microtype}

\usepackage{listings} 
\usepackage{graphicx}
\usepackage{caption}
\usepackage{subcaption} 

\usepackage[utf8]{inputenc}
\usepackage[T1]{fontenc}
\usepackage{hyphenat}
\usepackage{xspace}
\usepackage{amsmath}
\usepackage{amsfonts}
\usepackage{hyperref}
\usepackage{url}
\usepackage{booktabs}
\usepackage{multirow}
\usepackage{makecell}
\usepackage{caption}
\usepackage{minibox}
\usepackage{bbm}
\usepackage{graphicx}
\usepackage{balance}
\usepackage{mathtools}
\usepackage{color}
\usepackage{marvosym}
\usepackage{ifthen}
\usepackage{textcomp}
\usepackage{enumitem}
\usepackage{verbatim}
\usepackage{algorithm}
\usepackage{numprint}
\usepackage{balance}

\usepackage{amsthm}
\theoremstyle{plain}

\newcommand{\chatoDisplayMode}[1]{#1}

\definecolor{MyRed}{rgb}{0.6,0.0,0.0} 
\definecolor{MyBlack}{rgb}{0.1,0.1,0.1} 
\newcommand{\inred}[1]{{\color{MyRed}\sf\textbf{\textsc{#1}}}}
\newcommand{\frameit}[2]{
  \begin{center}
  {\color{MyRed}
  \framebox[.9\columnwidth][l]{
    \begin{minipage}{.85\columnwidth}
    \inred{#1}: {\sf\color{MyBlack}#2}
    \end{minipage}
  }\\
  }
  \end{center}
}

\newcommand{\note}[2][]{\chatoDisplayMode{\def\@tmpsig{#1}\frameit{{\Pointinghand} Note}{#2\ifx \@tmpsig \@empty \else \mbox{ --\em #1}\fi}}}
\newcommand{\todo}[2][]{\chatoDisplayMode{\def\@tmpsig{#1}\frameit{{\Writinghand} To-do}{#2\ifx \@tmpsig \@empty \else \mbox{ --\em #1}\fi}}}

\newcommand{\abbrevStyle}[1]{#1}

\newcommand{\ie}{\abbrevStyle{i.e.}\xspace}
\newcommand{\eg}{\abbrevStyle{e.g.}\xspace}
\newcommand{\cf}{\abbrevStyle{cf.}\xspace}

\newcommand{\Secref}[1]{Sec.~\ref{#1}}

\newcommand{\Tabref}[1]{Table~\ref{#1}}
\newcommand{\Figref}[1]{Fig.~\ref{#1}}

\newcommand{\xhdr}[1]{\vspace{1.7mm}\noindent{{\bf #1.}}}

\newcommand{\denselist}{ \itemsep -2pt\topsep-10pt\partopsep-10pt }

\newcommand{\textcite}[1]{\citeauthor{#1} \shortcite{#1}}

\newcommand{\hide}[1]{}

\makeatletter
\newcommand{\iffont}[2]{\ifthenelse{\equal{\f@family}{#1}}{#2}{}}
\makeatother

  \usepackage{mathptmx}

  \DeclareSymbolFont{greek}{OML}{cmm}{m}{n}
  \DeclareMathSymbol{\alpha}{\mathalpha}{greek}{"0B}
  \DeclareMathSymbol{\beta}{\mathalpha}{greek}{"0C}
  \DeclareMathSymbol{\gamma}{\mathalpha}{greek}{"0D}
  \DeclareMathSymbol{\delta}{\mathalpha}{greek}{"0E}
  \DeclareMathSymbol{\epsilon}{\mathalpha}{greek}{"0F}
  \DeclareMathSymbol{\zeta}{\mathalpha}{greek}{"10}
  \DeclareMathSymbol{\eta}{\mathalpha}{greek}{"11}
  \DeclareMathSymbol{\theta}{\mathalpha}{greek}{"12}
  \DeclareMathSymbol{\iota}{\mathalpha}{greek}{"13}
  \DeclareMathSymbol{\kappa}{\mathalpha}{greek}{"14}
  \DeclareMathSymbol{\lambda}{\mathalpha}{greek}{"15}
  \DeclareMathSymbol{\mu}{\mathalpha}{greek}{"16}
  \DeclareMathSymbol{\nu}{\mathalpha}{greek}{"17}
  \DeclareMathSymbol{\xi}{\mathalpha}{greek}{"18}
  \DeclareMathSymbol{\pi}{\mathalpha}{greek}{"19}
  \DeclareMathSymbol{\rho}{\mathalpha}{greek}{"1A}
  \DeclareMathSymbol{\sigma}{\mathalpha}{greek}{"1B}
  \DeclareMathSymbol{\tau}{\mathalpha}{greek}{"1C}
  \DeclareMathSymbol{\upsilon}{\mathalpha}{greek}{"1D}
  \DeclareMathSymbol{\phi}{\mathalpha}{greek}{"1E}
  \DeclareMathSymbol{\chi}{\mathalpha}{greek}{"1F}
  \DeclareMathSymbol{\psi}{\mathalpha}{greek}{"20}
  \DeclareMathSymbol{\omega}{\mathalpha}{greek}{"21}
  \DeclareMathSymbol{\varepsilon}{\mathalpha}{greek}{"22}
  \DeclareMathSymbol{\vartheta}{\mathalpha}{greek}{"23}
  \DeclareMathSymbol{\varpi}{\mathalpha}{greek}{"24}
  \DeclareMathSymbol{\varrho}{\mathalpha}{greek}{"25}
  \DeclareMathSymbol{\varsigma}{\mathalpha}{greek}{"26}
  \DeclareMathSymbol{\varphi}{\mathalpha}{greek}{"27}
  \DeclareSymbolFont{otone}{OT1}{cmr}{m}{n}
  \DeclareMathSymbol{\Gamma}{\mathalpha}{otone}{0}
  \DeclareMathSymbol{\Delta}{\mathalpha}{otone}{1}
  \DeclareMathSymbol{\Theta}{\mathalpha}{otone}{2}
  \DeclareMathSymbol{\Lambda}{\mathalpha}{otone}{3}
  \DeclareMathSymbol{\Xi}{\mathalpha}{otone}{4}
  \DeclareMathSymbol{\Pi}{\mathalpha}{otone}{5}
  \DeclareMathSymbol{\Sigma}{\mathalpha}{otone}{6}
  \DeclareMathSymbol{\Upsilon}{\mathalpha}{otone}{7}
  \DeclareMathSymbol{\Phi}{\mathalpha}{otone}{8}
  \DeclareMathSymbol{\Psi}{\mathalpha}{otone}{9}
  \DeclareMathSymbol{\Omega}{\mathalpha}{otone}{10}
  \DeclareSymbolFont{syms}{OML}{cmm}{m}{it}
  \DeclareMathSymbol{\partial}{\mathord}{syms}{"40}
  \DeclareMathAlphabet{\mathbold}{OML}{cmm}{b}{it}
  \DeclareSymbolFont{largesymbols}{OMX}{cmex}{m}{n}

\usepackage{algpseudocode} 
\usepackage{placeins}
\usepackage{float}
\usepackage{tabularx}


\title{GenIE: Generative Information Extraction}

\author{
Martin Josifoski{\normalfont ,}\textsuperscript{1}
Nicola De Cao{\normalfont ,}\textsuperscript{2,3}
Maxime Peyrard{\normalfont ,}\textsuperscript{1}
Fabio Petroni{\normalfont ,}\textsuperscript{4}
Robert West\textsuperscript{1}
\\
\textsuperscript{1}Ecole Polytechnique F\'ed\'erale de Lausanne \\
\textsuperscript{2}University of Amsterdam,
\textsuperscript{3}University of Edinburgh, \textsuperscript{4}Meta AI \\
\href{mailto:martin.josifoski@epfl.ch}{\texttt{{\color{black} martin.josifoski@epfl.ch}}},
\href{mailto:nicola.decao@gmail.com}{\texttt{{\color{black} nicola.decao@gmail.com}}} \\
\href{mailto:maxime.peyrard@epfl.ch}{\texttt{{\color{black} maxime.peyrard@epfl.ch}}}, 
\href{mailto:fabiopetroni@fb.com}{\texttt{{\color{black} fabiopetroni@fb.com}}},
\href{mailto:robert.west@epfl.ch}{\texttt{{\color{black} robert.west@epfl.ch}}}
}

\begin{document}
\maketitle

\begin{abstract}
Structured and grounded 
representation of
text is typically formalized by \emph{closed information extraction}, the problem of extracting an exhaustive set of \emph{(subject, relation, object)} triplets that are consistent with a predefined set of entities and relations from a knowledge base schema.
Most existing works are pipelines prone to error accumulation, and all approaches are only applicable to unrealistically small numbers of entities and relations. We introduce GenIE (generative information extraction), the first end-to-end autoregressive formulation of closed information extraction. 
GenIE naturally exploits the language knowledge from the pre-trained transformer by autoregressively generating relations and entities in textual form.
Thanks to a new bi-level constrained generation strategy, only triplets consistent with the predefined knowledge base schema are produced.
Our experiments show that GenIE is state-of-the-art on closed information extraction, generalizes from fewer training data points than baselines, and scales to a previously unmanageable number of entities and relations. 
With this work, closed information extraction becomes practical in realistic scenarios, providing new opportunities for downstream tasks. 
Finally, this work paves the way towards a unified end-to-end approach to the core tasks of information extraction.
\end{abstract}

\section{Introduction}
\label{sec:introduction}
\begin{figure}[ht!]
    \centering
    \includegraphics[scale=0.8]{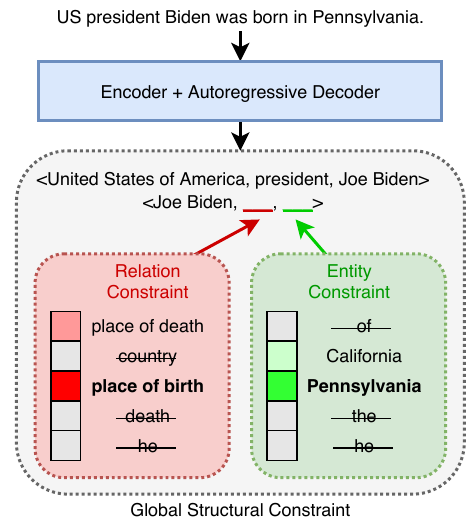}
    \caption{\textbf{Overview of GenIE.} We use a transformer encoder-decoder model that takes unstructured text as input and autoregressively generates a structured semantic representation of the information expressed in it, in the form of $($subject, relation, object$)$ triplets. 
    GenIE employs constrained beam search with: (i) a high-level constraint which asserts that the output corresponds to a set of triplets; (ii) lower-level constraints which use prefix tries to force the model to only generate valid entity or relation identifiers (from a predefined schema).
    %
    }
    \label{fig:genie}
\end{figure}

The ability to extract structured semantic information from unstructured texts is crucial for many AI tasks such as knowledge discovery~\citep{ji-grishman-2011-knowledge, trisedya-etal-2019-neural}, knowledge maintenance~\citep{tang-etal-2019-learning}, symbolic representation, and reasoning~\citep{Ji2021ASO}.
The interface between free text and structured knowledge is formalized by \emph{knowledge base population}~\citep[KBP;][]{ji-grishman-2011-knowledge}, which proposes to represent the information contained in text using \emph{(subject, relation, object)} fact triplets.
In this work, we focus on closed information extraction (cIE), the problem of extracting exhaustive sets of fact triplets expressible under the relation and entity constraints defined by a Knowledge Base (KB) schema.

Traditionally, cIE was approached with pipelines that sequentially combine named entity recognition~\citep{NER}, entity linking~\citep{EntityLinking}, and relation extraction~\citep{RelExtract}. Entity linking and relation extraction serve as grounding steps, matching entities and relations to numerical identifiers in a KB, e.g., QIDs and PIDs for Wikidata.
Recently,~\citet{trisedya-etal-2019-neural} pointed out that such pipeline architectures suffer from the accumulation of errors and proposed an end-to-end alternative. 
Nevertheless, existing methods are still only practical for small schemas with unrealistically small numbers of relations and entities.

Alternatively, some works have focused on a simpler syntactic task: open information extraction (oIE), which produces free-form triplets from texts. 
In this setup, the entities and relations are not grounded in a KB and, usually, do not represent facts~\citep{gashteovski-etal-2020-aligning}. 
As oIE triplets contain only surface relations, they have ambiguous semantics, making them hard to use in downstream tasks~\citep{BroscheitGA17} if not first aligned with a KB~\citep{gashteovski-etal-2020-aligning}.
Since, in practice, oIE often consists of structured substring selection,
it has recently been framed as an end-to-end sequence-to-sequence problem with great success~\citep{huguet-cabot-navigli-2021-rebel-relation, dognin-etal-2021-regen}. Indeed, such autoregressive formulations can exploit 
the language knowledge already encoded in pre-trained transformers~\citep{devlin-etal-2019-bert}.
For example, some tokens can be more easily recognized as possible entities or relations thanks to the pre-training information. 

Inspired by recent successes in oIE, we propose the first autoregressive end-to-end formulation of cIE that scales to many entities and relations, making cIE practical for more realistic KB schemas (\ie schemas with millions of entities).\footnote{Note that current methods, due to the atomic classification, have high memory requirements, and suffer from performance deterioration as the number of entities or/and relations grows.}
We employ a sequence-to-sequence BART model~\citep{lewis-etal-2020-bart}, and exploit a novel bi-level constrained generation strategy operating on the
space of possible triplets (from a fixed schema induced by Wikidata) to ensure that only valid triplets are generated.
Our resulting model, \textit{GenIE}, performs \textit{Gen}erative \textit{I}nformation \textit{E}xtraction and combines the advantages of a known schema with an autoregressive formulation. 
The high-level overview of GenIE is provided in \Figref{fig:genie}. The constrained generation encodes the known schema and enables the autoregressive decoder to generate textual tokens but only from the set of allowed entities or relations.

\xhdr{Contributions}
\begin{itemize}[topsep=0pt]
\denselist
    \item We present the first end-to-end autoregressive formulation of closed information extraction. 
    \item We describe a constrained decoding strategy that exploits the Wikidata schema to generate only valid fact triplets, demonstrating how constrained beam search can be applied on large, structured, and compositional spaces. 
    \item We propose a model that achieves state-of-the-art performance on the cIE task and scales to previously unmanageable numbers of entities (6M) and relations (more than 800).
    \item We point out and address weaknesses in the evaluation methodologies of recent previous works stemming from their small scale and the large imbalances in the available data per relation. We demonstrate the importance of reporting performance as a function of the number of relation occurrences in the data.
    \item We release pre-processed data, pre-trained models, and code within a general template designed to facilitate future research 
    at \url{https://github.com/epfl-dlab/GenIE}.
\end{itemize}

\section{Background and Related Work}
\label{sec:background}
\subsection{Closed Information Extraction }

In this work, we address the task of closed information extraction (cIE), which aims to extract the exhaustive set of facts from natural language, expressible under the relation and entity constraints defined by a knowledge base (KB). 

Most of the existing methods address the problem with a pipeline solution.
One line of work starts by first extracting the entity mentions and the relations between them from raw text. This is followed by a disambiguation step in which the entity and relation predicates are mapped to their corresponding items in the KB. 
The sub-task of extracting the free-form triplets was originally proposed by~\citet{etzioni2008open}, and it
is commonly referred to as open information extraction (oIE) or text-to-graph in the literature~\citep{guo-etal-2020-cyclegt, castro-ferreira-etal-2020-2020, huguet-cabot-navigli-2021-rebel-relation, EntLink}.
Another line of work employs a pipeline of models for (i) named entity recognition (NER) -- detecting the entity mentions; (ii) entity linking (EL) -- mapping the mentions to specific entities from the KB; (iii) relation classification (RC) -- detecting the relations that are expressed between the entities~\citep{galarraga2014canonicalizing, BootstrapSelf, chaganty-etal-2017-importance}. 
Due to their architecture, pipeline methods are plagued by error propagation, which significantly affects their performance~\citep{mesquita-etal-2019-knowledgenet, trisedya-etal-2019-neural}.

End-to-end systems that jointly perform the extraction and the disambiguation of entities and relations have been proposed to address the error propagation~\citep{trisedya-etal-2019-neural, sui-etal-2021-set, seq2RDF}. 
To mitigate the propagation of errors, these systems are endowed with the ability to leverage entity information in the relation extraction and vice-versa, which has resulted in significant performance gains. 
Conceptually, for producing the output triplets, existing methods all rely on atomic, multi-class classification-based ranking of relations and entities. Classification methods particularly suffer from imbalances in the data. On the contrary, our proposition, GenIE, is autoregressive and deals better with imbalances.

While cIE requires the constituent elements of the output triplets to be entities and relations associated with the KB, the output triplets in oIE are free-text. This makes the cIE task fundamentally harder than oIE and renders the majority, if not all, oIE methods inapplicable to the cIE setting. We report an additional discussion on relevant, but not fundamental, related work on oIE in Appendix~\ref{app:additional_related_work}.

\subsection{Autoregressive Entity Linking}

The tasks of entity linking (EL) and entity disambiguation (ED) have been extensively studied in the past~\citep{huang2015leveraging, wu-etal-2020-scalable, le-titov-2018-improving, kolitsas-etal-2018-end, arora-etal-2021-low}. 
Most existing approaches associate entities with unique atomic labels and cast the retrieval problem as multi-class classification across them. 
The match between the context and the label can then be represented as the dot product between the dense vector encodings of the input and the entity’s meta information~\citep{wu-etal-2020-scalable}.
This general approach has led to large performance gains.

Recently,~\citet{de-cao-etal-2021-highly,cao2021autoregressive,de2021multilingual} have suggested that the classification-based paradigm for retrieval comes with several shortcomings such as (i) the failure to capture fine-grained interactions between the context and the entities; (ii) the necessity of tuning an appropriately hard set of negative samples during training. 
Building on these observations, they propose an alternative solution that casts the entity retrieval problem as one of autoregressive generation in which the entity names are generated token-by-token in an autoregressive fashion. 
The (freely) generated output will not always be a valid entity name, and to solve this problem~\citet{cao2021autoregressive} propose a constrained decoding strategy that enforces this by employing a prefix trie.
Their method scales to millions of entities, achieving state-of-the-art performance on monolingual and multilingual entity linking.

Inspired by the intuition that language models are well suited for predicting entities, we propose a novel approach for cIE by framing the problem in an autoregressive generative formulation.

\section{Method}
\label{sec:method}
In this section we formalize GenIE, an autoregressive end-to-end model for closed information extraction. Let us assume a knowledge base (KB) consisting of a collection of entities $\mathcal{E}$, a collection of relations $\mathcal{R}$, and a set of facts
$(s, r, o) \in \mathcal{E} \times \mathcal{R} \times \mathcal{E}$
stored as (subject, relation, object) triplets. Additionally, we assume that each entity $e \in \mathcal{E}$ and relation $r \in \mathcal{R}$ is assigned to a textual label (corresponding to its name). The Wikidata KB~\citep{vrandevcic2012wikidata}, with Wikipedia page titles as entity names, and the Wikidata relation labels as relation names satisfy these assumptions. 

\subsection{Model}

We cast the task of information extraction as one of autoregressive generation. 
More concretely, given some text input $x$, GenIE strives to generate the linearized sequence representation $y$ of the exhaustive set of facts expressed in $x$.
The conditional probability (parameterized by $\theta$) assigned to the output $y$ is computed in the autoregressive formulation: $p_\theta(y\;|\;x) = \prod_{i=1}^{|y|} p_\theta(y_i\;|\;y_{<i}, x)$.
This can be seen as translating the unstructured text to a structured, unambiguous representation in a sequence-to-sequence formulation. GenIE employs the BART~\citep{lewis-etal-2020-bart} transformer architecture. It is trained to maximize the target sequence's conditional log-likelihood with teacher forcing~\citep{sutskever2011generating, sutskever2014sequence}, using the cross-entropy loss. We use dropout~\citep{JMLR:v15:srivastava14a} and label smoothing for regularization~\citep{szegedy2016rethinking}.

\subsection{Output Linearization}
\label{sec:linearization}

%
To represent the output with a sequence of symbols that is compatible with sequence-to-sequence architectures, we introduce the special tokens $\text{<sub>}$, $\text{<rel>}$, $\text{<obj>}$ to demarcate the start of the subject entity, the relation type and the object entity for each triplet. 
The special token $\text{<et>}$ is introduced to demarcate the end of the object entity, which is also the end of the triplet. 
We construct the sequence representation by concatenating the textual representations of its constituent triplets.
While the sequence representation has an intrinsic notion of order, the output set of triplets does not. 
To mitigate the effects of this discrepancy, we enforce a consistent ordering of the target triplets during training. 
Concretely, whenever the triplets' entities are linked to the entity mentioned in the textual input, we consider first the triplets for which the subject entity appears earlier in the text. 
Ties are resolved by considering the appearance position of the object entity. 

\subsection{Inference with Constrained Beam Search}
\label{ssec:inference}

The space of triplets corresponds to $\mathcal{T} = \mathcal{E} \times \mathcal{R} \times \mathcal{E}$, and the target space, which consists of triplet sets of arbitrary cardinality, is equivalent to $\mathcal{S} = \bigcup_{i=0}^{\infty} [\mathcal{E} \times \mathcal{R} \times \mathcal{E}]^i$.
At inference time, GenIE tackles the task of retrieving the linearized representation $y_S \in \mathcal{S}$ of a set of facts $S=\{t_1,\dots,t_n\}$ constituted by triplets $t_i \in \mathcal{T}$ expressed in the input text $x$.
Ideally, we would consider every element $y \in \mathcal{S}$ in the target space, assign it a score $p_\theta(y\;|\;x)$, and retrieve the most probable $y$.
Unfortunately, this is prohibitively expensive since we are dealing with a compositional target space whose size is gigantic (\eg, if we consider a Wikidata entity catalog of $|\mathcal{E}| \approx6\text{M}$ elements and a relation catalog of $|\mathcal{R}|\approx1000$ relations, that can express a total of $|\mathcal{T}|\approx10^{15}$ triplets; even if we limit ourselves to sentences that express only two facts, this provides us with $\approx10^{30}$ different output options).

On the other hand, the output needs to follow a particular structure, and contain only valid entity and relation identifiers. This does not necessarily hold for an arbitrary generation from a sequence-to-sequence model.

GenIE employs constrained beam search~\citep[BS;][]{sutskever2014sequence, cao2021autoregressive}
to resolve both of these problems.
Instead of explicitly scoring all of the elements in the target space $\mathcal{S}$, the idea is to search for the top-$k$ eligible options, using BS with $k$ beams and a prefix trie. BS considers one step ahead – the next token to be generated – conditioned on the previous ones. 
The prefix trie restricts the BS to candidate tokens that could lead to valid identifiers.
However, for the cIE setting we are interested in, the target space is prohibitively large to pre-compute the necessary trie. Therefore, we enforce a bi-level constraint on the output that allows for compositional, dynamic generation of the valid prefixes. More specifically, GenIE enforces: (i) a high-level structural constraint which asserts that the output follows the linearization schema defined in Sec.~\ref{sec:linearization}; (ii) lower level validity constraints which use an entity trie and a relation trie to force the model to only generate valid entity or relation identifiers, \mbox{respectively -- depending} on the specific element of the structure that is being generated. This outlines a general approach for applying BS to search through large compositional structured spaces.












\section{Experimental Setup}
\label{sec:experiments_setup}
\subsection{Knowledge Base: Wikidata}
We use Wikidata\footnote{Dumps from 2019/08/01}~\citep{vrandevcic2012wikidata} as the target KB to link to, filtering out all entities that do not have an English Wikipedia page associated with them. The filtering guarantees that all entity names are unique.
Our final entity set $\mathcal{E}$ contains 5,891,959 items. We also define our relation set $\mathcal{R}$ as the union of all the relations considered in the datasets described below, resulting in 857 relations. For different datasets, we consider only the subset of annotated relations to better compare with baselines.
Although large, the number of entity (and relation) names is not a memory bottleneck as the generated prefix trie occupies $\approx$200MB of storage (\eg, the entity linking system proposed by \citealt{wu-etal-2020-scalable} needs $>$20 times more storage).

\subsection{Datasets and Evaluation Metrics}
In this work, we further annotate and adapt REBEL~\citep{huguet-cabot-navigli-2021-rebel-relation} and
Wiki-NRE for training, validation and testing. Additionally, we use Geo-NRE~\citep{trisedya-etal-2019-neural}, and FewRel~\citep{han-etal-2018-fewrel} for testing purposes only. Appendix~\ref{app:datasets} contains descriptions of these datasets and their statistics. 
We measure the performance in terms of micro and macro precision, recall and F1.
See Appendix~\ref{sec:perf_metrics} for a detailed and formal description of these metrics. We also report a 1-standard-deviation confidence interval constructed from 50 bootstrap samples of the data.



\subsection{Baselines}
\label{sec:baselines}

We compare GenIE against Set Generation Networks~\citep[SetGenNet;][]{sui-etal-2021-set} which is, to the best of our knowledge, the strongest 
model on Wiki-NRE and Geo-NRE.
Note that the authors did not release code or the model and there is no other model from the literature trained and evaluated on REBEL for cIE. SetGenNet~\citep{sui-etal-2021-set} is an end-to-end state-of-the-art model for triplet extraction. It consists of a transformer encoder~\citep{vaswani2017attention} that encodes the input followed by a non-autoregressive transformer decoder~\citep{gu2018nonautoregressive}. The decoder generates embeddings that are used to predict entities and relations. SetGenNet further uses candidate selection~\citep{ganea-hofmann-2017-deep, kolitsas-etal-2018-end} to reduce the output space and a bipartite matching loss that handles different prediction orderings (\ie, it generates a set). Note that there are weaker baselines (e.g., \citealt{trisedya-etal-2019-neural}) we could have used to compare on REBEL, but we were not able to reproduce their code. We report details on the effort made to use these baselines in Appendix~\ref{app:baselines}.

We also implement a pipeline baseline, consisting of 4 independent steps, namely: (i) \textit{named entity recognition} (NER), which selects the spans in the input source likely to be entity mentions; (ii) \textit{entity disambiguation} (ED), which links mentions to their corresponding identifiers in the KB; (iii) \textit{relation classification} (RC), which predicts the relation between a given pair of entities, and finally; (iv) \textit{triplet classification} (TC), which predicts whether a given triplet is actually entailed by the context.
TC is necessary because the previous step (RC) predicts a relation for every pair of entities. Each step needs to be trained independently with a specific architecture tailored for the task, and we made an optimal choice for each step. For the NER component we used the state-of-the-art tagger FLAIR\footnote{\url{https://github.com/flairNLP/flair}}~\citep{akbik2019flair}, while for ED we used the GENRE linker\footnote{\url{https://github.com/facebookresearch/GENRE}}~\citep{cao2021autoregressive}. These two models were already trained, and we use them for inference only. For RC and TC, we trained a RoBERTa~\citep{liu2019roberta} model with a linear classification layer on top (as these two sub-tasks are typically cast as classification problems). \citet{trisedya-etal-2019-neural} also proposed many other pipeline baselines but ours 
outperforms
them (see \Tabref{tab:baseline_results} in Appendix~\ref{app:additional_results} for comparison).

\section{Results}
\label{sec:experiments_results}

\subsection{Performance Evaluation}
\label{ssec:perf_eval}

\begin{table*}[t]
\centering
\resizebox{\textwidth}{!}{
\setlength{\tabcolsep}{5pt}
\begin{tabular}{@{}lccc|ccc|ccc|c@{}}
\toprule
& \multicolumn{6}{c}{\emph{\textbf{Small Evaluation Schema}}} & \multicolumn{4}{c}{\emph{\textbf{Large Evaluation Schema}}} \\
& \multicolumn{3}{c}{Wiki-NRE} & \multicolumn{3}{c}{Geo-NRE} & \multicolumn{3}{c}{REBEL} & FewRel \\
& Precision & Recall & F1 & Precision & Recall & F1 & Precision & Recall & F1 & Recall \\
\midrule
\midrule
\multicolumn{3}{l}{\emph{\textbf{Micro}}} \\ 
\hspace{4mm} SetGenNet (W)          &  82.75 {\scriptsize± 0.11} &  77.55 {\scriptsize± 0.27} & 80.07 {\scriptsize± 0.27} &  86.89 {\scriptsize± 0.51} & 85.31 {\scriptsize± 0.47} & 86.10 {\scriptsize± 0.34} &  -- & -- & --     &  -- \\
\hspace{4mm} SotA Pipeline (W)    &    67.43 {\scriptsize± 0.28}    &    54.22 {\scriptsize± 0.21}    &    60.11 {\scriptsize± 0.22}    &    64.60 {\scriptsize± 1.46}    &    64.05 {\scriptsize± 1.46}    &    64.32 {\scriptsize± 1.45} &  -- & -- & --     &  -- \\
\hspace{4mm} SotA Pipeline (R)    &    50.78 {\scriptsize± 0.20}    &    62.17 {\scriptsize± 0.24}    &    55.90 {\scriptsize± 0.20}    &    60.28 {\scriptsize± 1.45}    &    60.78 {\scriptsize± 1.49}    &    60.53 {\scriptsize± 1.45} &  43.30 {\scriptsize± 0.15} & 41.73 {\scriptsize± 0.13}  & 42.50 {\scriptsize± 0.13}     &  17.89 {\scriptsize± 0.24} \\
\hspace{4mm} SotA Pipeline (R+W)    &    65.17 {\scriptsize± 0.27}    &    54.40 {\scriptsize± 0.20}    &    59.30 {\scriptsize± 0.21}    &    66.65 {\scriptsize± 1.47}    &    66.22 {\scriptsize± 1.46}    &    66.43 {\scriptsize± 1.45} &  -- & -- & --     &  -- \\
\midrule
\hspace{4mm} GenIE (W)                   &  88.18 {\scriptsize± 0.13} & 88.31 {\scriptsize± 0.16} & 88.24 {\scriptsize± 0.13}     &  86.46 {\scriptsize± 1.05} & 87.14 {\scriptsize± 1.03} & 86.80 {\scriptsize± 1.03} &  -- & -- & --     &  -- \\
\hspace{4mm} GenIE (R)                   &  27.98 {\scriptsize± 0.13} & 67.16 {\scriptsize± 0.20} & 39.50 {\scriptsize± 0.14}     &  39.69 {\scriptsize± 1.65} & 59.01 {\scriptsize± 1.56} & 47.45 {\scriptsize± 1.62} &  68.02 {\scriptsize± 0.15} & 69.87 {\scriptsize± 0.14} & 68.93 {\scriptsize± 0.12}     &  30.77 {\scriptsize± 0.27} \\
\hspace{4mm} GenIE (R+W)             &  \textbf{91.39 {\scriptsize± 0.15}} & \textbf{91.58 {\scriptsize± 0.14}} & \textbf{91.48 {\scriptsize± 0.12}}    & \textbf{91.77 {\scriptsize± 0.98}} & \textbf{93.20 {\scriptsize± 0.83}} & \textbf{92.48 {\scriptsize± 0.88}} &  -- & -- & --     &  -- \\

\midrule                            
\midrule                            
\multicolumn{3}{l}{\emph{\textbf{Macro}}} \\
\hspace{4mm} SotA Pipeline (W)    &    11.96 {\scriptsize± 0.72}    &    10.73 {\scriptsize± 0.46}    &    10.56 {\scriptsize± 0.43}    &    24.82 {\scriptsize± 3.61}    &    22.54 {\scriptsize± 3.67}    &    20.39 {\scriptsize± 2.72} &  -- & -- & --     &  -- \\
\hspace{4mm} SotA Pipeline (R)    &    19.39 {\scriptsize± 1.18}    &    17.41 {\scriptsize± 0.99}    &    15.93 {\scriptsize± 0.93}    &    28.80 {\scriptsize± 3.86}    &    30.24 {\scriptsize± 4.46}    &    25.24 {\scriptsize± 3.21} &  12.20 {\scriptsize± 0.35} & 10.44 {\scriptsize± 0.22}  & 9.48 {\scriptsize± 0.21}     &  19.67 {\scriptsize± 0.26} \\
\hspace{4mm} SotA Pipeline (R+W)    &    24.12 {\scriptsize± 1.46}    &    16.55 {\scriptsize± 1.00}    &    17.76 {\scriptsize± 1.01}    &    38.67 {\scriptsize± 5.72}    &    34.49 {\scriptsize± 5.99}    &    35.14 {\scriptsize± 5.09} &  -- & -- & --     &  -- \\
\midrule
\hspace{4mm} GenIE (W)                   &  44.22 {\scriptsize± 2.40} & 36.79 {\scriptsize± 1.62} & 38.39 {\scriptsize± 1.71}     &  57.13 {\scriptsize± 6.83} & 52.83 {\scriptsize± 6.84} & 52.79 {\scriptsize± 6.27} &  -- & -- & --     &  -- \\
\hspace{4mm} GenIE (R)                  &  30.63 {\scriptsize± 1.40} & 41.97 {\scriptsize± 1.92} & 29.27 {\scriptsize± 1.26}     &  32.38 {\scriptsize± 5.86} & 40.39 {\scriptsize± 5.17} & 30.67 {\scriptsize± 5.23} &  33.90 {\scriptsize± 0.73} & 30.48 {\scriptsize± 0.65} & 30.46 {\scriptsize± 0.62}     &  30.78 {\scriptsize± 0.26} \\
\hspace{4mm} GenIE (R+W)             &  \textbf{52.55 {\scriptsize± 2.12}} & \textbf{45.95 {\scriptsize± 1.67}} & \textbf{47.08 {\scriptsize± 1.68}}     &  \textbf{75.77 {\scriptsize± 7.80}} & \textbf{71.60 {\scriptsize± 7.95}} & \textbf{72.59 {\scriptsize± 7.32}} &  -- & -- & --     &  -- \\
\bottomrule                            
\end{tabular}
}
\caption{\textbf{Main results.} ``R'' indicates training on REBEL, and ``W'' indicates training on Wiki-NRE.}
\vspace{-1em}
\label{tab:eval-systems}
\end{table*}


Models performing cIE can base their predictions on different schemas. 
In this section, we distinguish between \emph{small} and \emph{large evaluation schema}. The \emph{small evaluation schema} is consistent with previous approaches where models only have to decide between a small set of relations and entities (the schema induced by Wiki-NRE). In the \emph{large evaluation schema}, models use the schema induced by REBEL. Models also use the large evaluation schema of REBEL when tested on FewRel, as a high-quality and challenging recall-based evaluation.
We consider 3 training setups for GenIE and the pipeline baseline comprised of SotA components:
(i) the training set of Wiki-NRE (W) only, (ii) the training set of REBEL (R) only, and (iii) pre-training on REBEL and fine-tuning on Wiki-NRE (R+W). The implementation details are given in Appendix~\ref{app:implementation_details}.
%
We report the macro and micro precision, recall, and F1 in \Tabref{tab:eval-systems}. Unfortunately, as the code for SetGenNet is not available, we cannot compute its macro performance, thus we report the micro scores only.

\begin{table}
\centering
\resizebox{0.48\textwidth}{!}{
\setlength{\tabcolsep}{5pt}
\begin{tabular}{@{}lccc|c@{}}
\toprule
& \multicolumn{3}{c}{REBEL} & FewRel \\
& Precision & Recall & F1 & Recall \\
\midrule
\midrule
\emph{\textbf{Micro}} \\ 
\hspace{4mm} GenIE    &    \textbf{68.02 {\scriptsize± 0.15}}    &    69.87 {\scriptsize± 0.14}    &    68.93 {\scriptsize± 0.12}    &    30.77 {\scriptsize± 0.27} \\
\hspace{4mm} GenIE - PLM    &    59.32 {\scriptsize± 0.13}    &    \textbf{77.78} {\scriptsize± 0.12}    &    67.31 {\scriptsize± 0.10}    &    \textbf{46.95} {\scriptsize± 0.27} \\
\hspace{4mm} GenIE - GENRE    &    64.14 {\scriptsize± 0.14}    &    76.58 {\scriptsize± 0.11}    &    \textbf{69.81 {\scriptsize± 0.10}}    &    \textbf{46.62} {\scriptsize± 0.25} \\
\hspace{4mm} GenIE unconstrained    &    65.30 {\scriptsize± 0.14}    &    67.12 {\scriptsize± 0.12}    &    66.20 {\scriptsize± 0.11}    &    26.15 {\scriptsize± 0.27} \\
\midrule
\emph{\textbf{Macro}} \\ 
\hspace{4mm} GenIE    &    \textbf{33.90 {\scriptsize± 0.73}}    &    30.48 {\scriptsize± 0.65}    &    30.46 {\scriptsize± 0.62}    &    30.78 {\scriptsize± 0.26} \\
\hspace{4mm} GenIE - PLM    &    30.66 {\scriptsize± 0.68}    &    \textbf{43.33} {\scriptsize± 0.63}    &    \textbf{33.85} {\scriptsize± 0.58}    &    \textbf{46.96} {\scriptsize± 0.25} \\
\hspace{4mm} GenIE - GENRE    &    32.02 {\scriptsize± 0.67}    &    39.14 {\scriptsize± 0.68}    &    \textbf{33.40} {\scriptsize± 0.62}    &    \textbf{46.63} {\scriptsize± 0.24} \\
\hspace{4mm} GenIE unconstrained    &    32.25 {\scriptsize± 0.66}    &    27.59 {\scriptsize± 0.53}    &    28.20 {\scriptsize± 0.50}    &    26.14 {\scriptsize± 0.24} \\
\bottomrule                            
\end{tabular}
}
\caption{\textbf{Ablation study on the weights initialization and the constrained generation strategy.}}
\label{tab:ablation}
\end{table}

First, we observe a large and significant F1 improvement of $8$ and $28$ absolute points obtained by GenIE over SetGenNet and the pipeline baseline, respectively, when trained on the same dataset (W). 
Despite the much bigger schema employed by REBEL, pre-training on it and then fine-tuning (R+W), improves the performance on Wiki-NRE and Geo-NRE for $3\%$ and $5\%$, respectively. This highlights that: (i) GenIE can effectively transfer knowledge across datasets/schemas; (ii) GenIE can quickly adapt to new schemas. Due to its rigid, monolithic relation classifier, the pipeline baseline does not possess these qualities. However, the pre-training does improve its macro scores.

Only the newly developed pipeline baseline and GenIE can scale-up to the larger schema\footnote{See notes on reproducibility in Appendix~\ref{app:baselines}.}, and as expected, this setting is more challenging for both models. However, GenIE still preserves a good F1 score of $68$ micro and $34$ macro, which is a relative increase of $60 \%$ and $320 \%$, respectively, over the baseline. While the pipeline has a steeper drop from micro to macro scores, in general, a significant difference between the two is observed in every setting. This suggests that the models perform better for relations associated with many training examples and significantly worse for the rest. These findings call for the fine-grained analysis of performance in \Secref{ssec:error_analysis} that partitions the relations according to their occurrence count in the training data. For completeness, we also provide an analysis of performance as a function of the number of relations considered, in Appendix~\ref{app_ssec:eval_n_rel}.

Finally, on FewRel, recall is the only well-defined metric (see Appendix~\ref{app:datasets}). In this setting as well, GenIE greatly outperforms the baseline by 13 (micro) and 11 (macro) recall points (micro and macro are close as the dataset is class-balanced).

\xhdr{Ablation study} In \Tabref{tab:ablation} we summarize the results of an ablation study considering the pre-training and the constrained generation. We consider three different starting points: (i) a random initialization; (ii) BART~\citep{lewis-etal-2020-bart} pre-trained language model (PLM); (iii) a pre-trained autoregressive entity retrieval model GENRE \citep{cao2021autoregressive}. The pre-trained models are better in terms of recall and exhibit a better out-of-domain generalization on FewRel. In contrast, they are slightly worse in terms of precision, which translates to maximum improvement of a single point in F1 on REBEL. Another salient advantage of pre-training is reducing the training steps necessary for achieving good results. Indeed, when starting from GENRE, 3-5k steps are sufficient for competitive performance; starting from a PLM necessitates 5-10k steps; while a random initialization requires 40-50k steps for competitive performance. 
Additionally, the pre-trained versions converge to a lower validation loss (see \Figref{fig:training_loss} in Appendix~\ref{app:additional_results}).

To quantify the benefits from the constrained generation, we compare the results attained by the randomly initialized model with and without constraints. In addition to ensuring a structure on the output, the constrained generation strategy results in an increase of $2$-$3$ absolute points in terms of F1.




\subsection{Analysis of Performance as a Function of the Relation Occurrence Count}
\label{ssec:error_analysis}

\begin{figure*}
    \centering
    \includegraphics[width=0.8\textwidth]{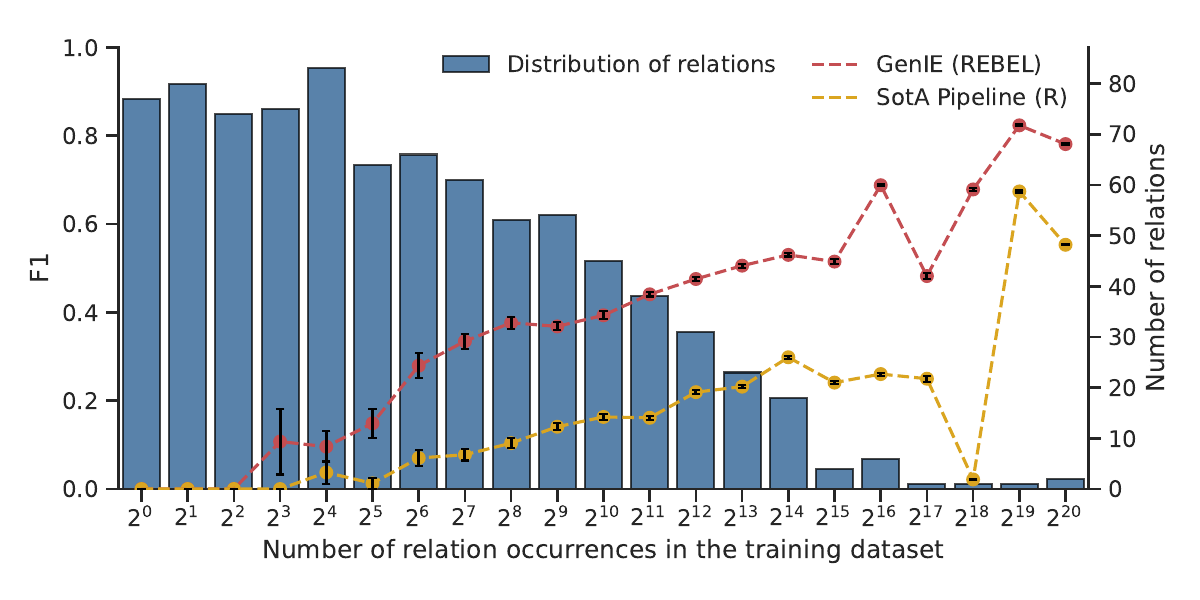}
    \vspace{-1em}
    \caption{\textbf{Impact of the number of relation occurrences.} Relations are bucketed based on their number of occurrences; bucket $2^i$ contains relations occurring between $2^{i}$ and $2^{i+1}$ times. The histogram shows the number of relations per bucket. The line plots depict the F1 scores of GenIE and the baseline per bucket together with confidence intervals computed per bucket with bootstrap resampling.}
    \label{fig:bucket_plot}
\end{figure*}

The datasets naturally present large imbalances, where few relations occur a lot, but most relations are rare.
In the previous section, we already observed a large difference between macro and micro F1 scores of models, indicating that the number of occurrences impacts model performances.
Thus, we now measure F1 scores after bucketing relations according to their number of occurrences in the training dataset. 
In \Figref{fig:bucket_plot}, we create buckets $i \in \{0 , \dots, 20\}$, where bucket $i$ contains all the relations occurring at least $2^i$ times and less than $2^{i+1}$ times in the REBEL dataset. 
The height of the histogram for bucket $i$ shows how many relations are contained in this bucket.
Finally, we report the F1 scores of GenIE and the pipeline of SotA components per bucket.
Note that micro F1 from \Tabref{tab:eval-systems} is equivalent to putting all relations in one single bucket (equal weight to each data point), and macro F1 is equivalent to averaging the F1 with one bucket per relation (equal weight to each relation).

The histogram first confirms that most of the relations occur in only a few triplets from the training data. Models thus need to perform few-shot learning for most of the relations.
GenIE is significantly better than the pipeline baseline for all the buckets. 
Finally, it is important to highlight that even though the performance of both methods, unsurprisingly, declines for relations that appear less often in the training data, GenIE already performs well for relations with at least $2^6=64$ occurrences. On the contrary, the baseline needs $2^{14}=16{,}384$ samples to reach a comparable level of performance, and scores better than GenIE does for the $2^6=64$ bucket only after seeing at least $2^{19}=524{,}288$ samples. 
This confirms that GenIE is not only better at macro and micro F1, but it is also capable of performing fewer-shot learning than the baseline.
It further shows that, contrary to the baseline, GenIE's good scores do not come solely from its ability to perform well on the few most frequent relations.

\subsection{Disentangling the Errors}
\label{ssec:disentangling}

The task of cIE, explicitly or implicitly, encompasses NER, NEL and RC as its subtasks. Failure in any of them directly translates to failure on the original task. Therefore, to effectively compare different cIE models and accurately characterize their behavior, we need to evaluate their performance on each of the subtasks. 

\begin{figure}
    \centering
    \includegraphics[width=\linewidth]{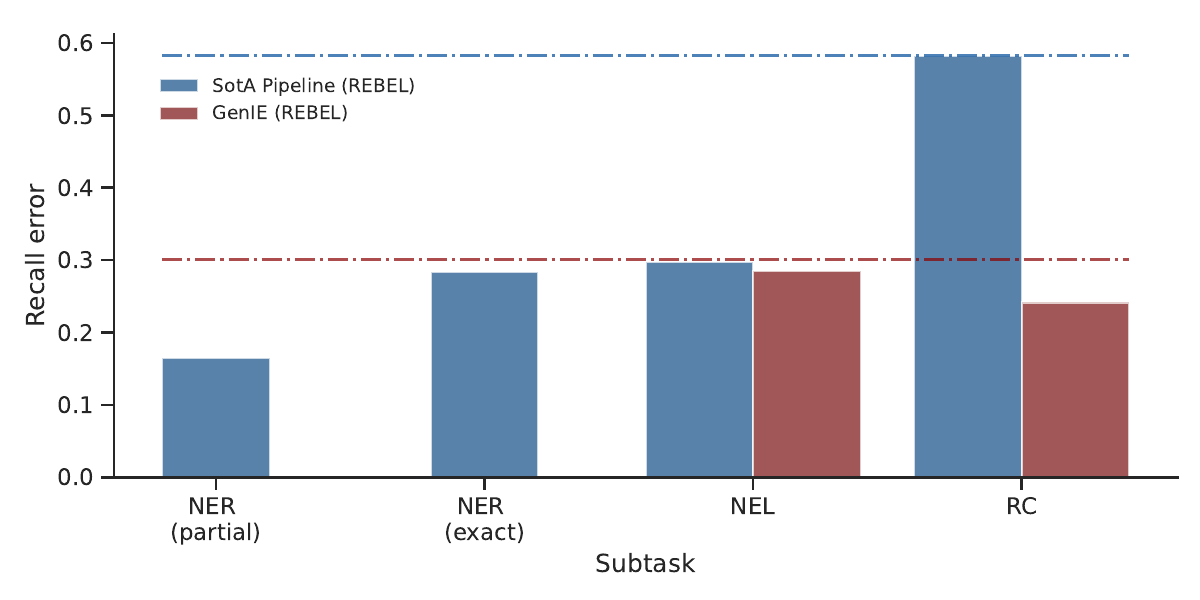} 
    \vspace{-22pt}
    \caption{\textbf{Attribution of error to each of the cIE subtasks.} The dashed lines equal the overall recall error of the system. Lower is better.}
    \label{fig:error_attribution}
\end{figure}
The separation of responsibility between the pipeline components leads to a natural error attribution for the SotA pipeline model. To estimate the NER error, we take the entity mentions predicted by the NER component and compare them with the corresponding mentions of the constituent entities of the output triplets. All triplets that concern an entity whose mention was not retrieved by the NER component are considered erroneous. We differentiate between two settings: (i) exact, which requires that the generated mention exactly matches the target mention; and (ii) partial, for which any overlap between the generated and the target triplet is sufficient. The NEL error is calculated by considering the output of the NEL component and comparing the linked entities to those in the output triplets. Again, all of the triplets that concern an incorrectly linked entity are considered erroneous. Finally, for the pipeline, every correctly predicted relation label translates to a correctly extracted triplet. Therefore, we calculate the RC error using the cIE definition of recall given in Appendix~\ref{sec:perf_metrics}.

End-to-end systems tackle all of the sub-tasks jointly, which makes the error attribution, in this setting, less obvious. To estimate errors corresponding to a particular target triplet, we need a reference triplet -- among the predicted ones -- for comparison. We start by outlining a bipartite matching procedure. Let each triplet be a node in a graph. We add an edge between each target--prediction pair of triplets. The edges are assigned a weight determined as a function of the pair of triplets they connect. Concretely, an edge $e$ that connects the target triplet $t_T = (s_T, r_T, o_T)$ and the predicted triplet $t_P = (s_P, r_P, o_P)$ will be assigned a weight of: \textit{1}, if the triplets are the same; \textit{2}, if they express the same relation, but either the subject or the object differ; \textit{3}, if they concern the same entities, but the relation differs; \textit{4}, if they share only a single entity; \textit{5}, if they share only a single relation; \textit{6} if they have nothing in common. We construct the matching by selecting edges in a greedy fashion until all of the target triplets have been matched. The procedure ensures that every target triplet is paired with its closest match. Finally, we estimate the NEL error as the portion of edges that were assigned a weight $w \in \{2, 4, 5, 6\}$, and the RC error as the portion of edges that where assigned a weight $w \in \{3, 4, 6\}$.

The results of this analysis are summarized in \Figref{fig:error_attribution}. Immediately, the NER component in the pipeline method introduces an 18\% error by completely missing on relevant entity mentions. An additional 12 absolute points hinge on a partial matching.
The NEL component matches most of the entity mentions that are retrieved, but at this point, even with a (hypothetical) perfect RC, the performance of the pipeline will be only on par with GenIE. 
In practice, the RC component adds 30\% to the inherited error, effectively doubling it.

On another note, the absolute error attributed to NEL by the pipeline and GenIE differs in a few absolute points only, while the difference for the (non-inherited) error stemming from RC is less than 10\%. Adding these two together leaves us much shorter than the actual gap of 28 absolute points in performance on the cIE task. 
This suggests a strong correlation between the performance on NEL and RC for GenIE, which is fueled by the increased flow of information between the subtasks. 
The previous allows for more fine-grained interactions between the entities, the relations, and the context to be captured, consequently improving the overall performance. 
Alternatively, whenever the model captures a misleading correlation/interaction, it is amplified and hinders the performance on both subtasks. 
This result is echoed by the fact that the sum of the errors attributed to NEL and RC is significantly smaller than the error on the cIE task. 
Based on this observation, we hypothesize that any improvement on NEL will overflow to the RC subtask – and vice-versa – thereby directly translating to performance gains on the overall task.

%
%

\section{Discussion}
\label{sec:discussion}

\xhdr{Unifying the cIE spectrum} 
There is a full spectrum of tasks that are closely related to cIE and are central to the field of information extraction.
The typical setup assumes a KB associated with entities and relations, and the goal is to either annotate the text with information from the KB, or extract structured unambiguous information from the text.
The tasks of entity linking and relation classification, already discussed in \Secref{sec:background} and \Secref{sec:baselines}, are two such examples. 
Another example is slot filling (SF), the task of extracting information for a specific entity and relation (\eg, entity \textit{Mick Jagger}, relation \textit{member of}) from natural language~\citep{Surdeanu2013OverviewOT, petroni-etal-2021-kilt}.

All of these problems rely on the same set of logical tasks: identifying entities from the KB in text and understanding how they interact. Therefore, it would be beneficial to assume a single model, or a set of models that share parts of the weights and collectively solve all of the tasks. 
This would allow for the information from a dataset collected for one task (\eg, RC) to be leveraged for improving the performance of another (\eg, SF or cIE).


\xhdr{Bridging the gap between oIE and cIE} 
In this work, we only considered triplets for which both entities are element in the entity catalog. However, for many useful relations one of the objects is a literal~\citep{mesquita-etal-2019-knowledgenet}, \eg, date of birth, length, size, number of employees or others. GenIE can be readily extended to accommodate for this, by adapting the decoding strategy allowing that for specific relations the entity can be a substring from the input.  This is subtle connection to oIE which has thus far been treated as a separate problem.
Current state-of-the-art methods on the oIE task address the problem in a similar autoregressive formulation (see Appendix~\ref{app:additional_related_work} for more discussion).

\xhdr{Real world implications}
%
%
%
Generative models have been shown to be very effective even in massive multilingual settings---e.g.,~\citet{de2021multilingual} proposed mGENRE, a multilingual version of GENRE trained and tested on more than 100 language. Our GenIE formulation would not need substantial modifications to adapt to such setting. 
Having a single model that works in hundreds of languages would be extremely useful and a very promising direction for future work.

While autoregressive models have a non-negligible computation footprint, \citet{de-cao-etal-2021-highly} show that autoregressive EL can be sped up 70x with no cost on performance.
The fact that this solution can be adapted to GenIE makes the practical impact of our method even greater.



\section{Conclusion}
\label{sec:ccl}
This paper provides a new view on closed information extraction (cIE) by casting the problem as autoregressive sequence-to-sequence generation. 
Our method, GenIE, leverages the autoregressive formulation to capture the fine-grained interactions expressed in the text and employs a bi-level constrained generation strategy to effectively retrieve the target representation from a large, structured, compositional predefined space of outputs.
Experiments show that GenIE achieves state-of-the-art performance on cIE and can scale to a previously unmanageable number of entities and relations.
We believe that our autoregressive formulation of cIE, coupled with constrained decoding, is a stepping stone towards a unified approach for addressing the core tasks in information extraction. 


\section*{Acknowledgments}
Nicola De Cao is supported by SAP Innovation Center Network. Authors want to thank Ivan Titov and Wilker Aziz for insightful discussions and help. With support from Swiss National Science Foundation (grant 200021\_185043), European Union (TAILOR, grant 952215), and gifts from Microsoft, Facebook, Google.

\clearpage
\bibliography{main}
\bibliographystyle{acl_natbib}

\appendix
\clearpage

\section{Additional Background and Related Work} \label{app:additional_related_work}

\subsection{Generative Open Information Extraction}

Early work had focused on pipeline architecture for oIE. In general, these methods first detect the entity mentions present in the text and then, for  pair of entities, in a classification setting, predict the existence of a relation between the two entities and the relation type~\citep{angeli-etal-2015-leveraging, del2013clausie}. 
The advent of transformers~\citep{devlin-etal-2019-bert, Lan2020ALBERT:, liu2019roberta} and pipeline architectures that allow for information to flow between the two subtasks – usually by sharing some parameters of the encoder – have allowed these models to do well on the oIE task. 
However, they do come with some general limitations: (i) assuming the existence of a single relation between a pair of entities; (ii) inability to capture the interactions between triplets.

Much of the current research is focused on studying the oIE problem in the autoregressive generative setting, which seamlessly mitigates the limitations mentioned above~\citep{huguet-cabot-navigli-2021-rebel-relation, dognin-etal-2021-regen, Nayak_Ng_2020}. For instance, ReGen~\citep{dognin-etal-2021-regen} significantly improves upon published results and establishes state-of-the-art results on the dataset used in the WebNLG 2020+ Challenge~\citep{castro-ferreira-etal-2020-2020}. REBEL~\citep{huguet-cabot-navigli-2021-rebel-relation}, on the other hand, achieves state-of-the-art performances across a suite of oIE benchmarks. Moreover, both of these methods address the problem in a similar formulation that takes the text as input context and generates the output triplets token-by-token in an autoregressive fashion.

The output triplets in oIE are free-text, while cIE requires the constituent elements of the output triplets to come from the entity and relation sets associated with the KB. This makes the cIE task fundamentally harder than oIE, and renders these methods not applicable to the cIE setting.

Finally,~\citet{taille-etal-2020-lets} make an effort to describe the many issues with the evaluation of oIE systems in literature and call for a unified evaluation setting for a fair comparison between systems. These problems get only exacerbated in cIE where the performance of a model would highly depend on the entity and relation catalogue considered. To alleviate some of these issues, we annotate the REBEL dataset~\citep{huguet-cabot-navigli-2021-rebel-relation} with unique textual entity identifiers and textual relation labels, and propose a suite of meaningful evaluation settings while considering an approximately  6 million long entity catalogue comprised of all the entities in the English Wikipedia, and 857 long relation catalogue supported by the dataset. 

\begin{table*}[t]
    \centering
    {
    \setlength{\tabcolsep}{5pt}
    \begin{tabular}{lrrr|rrr|rr}
        \toprule
        \textbf{Dataset} & \multicolumn{3}{c}{\textbf{Documents}}  & \multicolumn{3}{c}{\textbf{Triplets}} & $|\mathcal{E}|$\textsuperscript{\textdagger} & $|\mathcal{R}|$\textsuperscript{\textdagger} \\
        & training & validation & test & training & validation & test & \\
        \midrule
        REBEL & 1,899,331 & 104,960 & 105,516 & 5,147,836 & 284,268 & 284,936 & 1,498,143 & 857 \\
        Wiki-NRE & 223,536 & 980 & 29,619 & 298,489 & 1,317 & 39,678 &  278,204 & 157 \\
        Geo-NRE & -- & -- & 1,000 & -- & -- & 1,000 & 124 & 11 \\
        FewRel & -- & 26,892\textsuperscript{$\star$} & 27,650 & -- & 26,892\textsuperscript{$\star$} & 27,650 & 64,762 & 80 \\ 
        \bottomrule
    \end{tabular}
    }
    \caption{\textbf{Statistics of the datasets.} \textsuperscript{\textdagger}With an abuse of notation here we indicate the amount of unique entities and relations for each dataset and not the size of the Knowledge Base associated with it (see Section~\ref{sec:experiments_setup} for more details). \textsuperscript{$\star$} Note that we do not use the validation FewRel data in our experiment, but we release this split as well.}
    \label{tab:datasets}
\end{table*}

\section{Datasets} \label{app:datasets}
\Tabref{tab:datasets} summarizes the statistics of all datasets used in this work. 
For each dataset, we remove datapoints containing triplets with entities that do not have an associated Wikipedia page (\ie, entities not associated to a unique name).
This filtering removes a negligible portion of the data in most cases (\ie, $<$0.5\%) except for REBEL where 3.4\% of datapoints were removed.

We evaluate the models in a standard setups for Wiki-NRE and Geo-NRE. For these datasets, the schema is unrealistically small: $\approx$300K entities with 157 relations for Wiki-NRE and 124 entities with 11 relations for Geo-NRE. Therefore, we scale to previously unexplored schema sizes for cIE using the REBEL dataset ( $\approx$6M entities and 857 relations). We also use FewRel as a high-quality dataset for recall evaluation using the large schema from REBEL.

\paragraph{REBEL}\citep{huguet-cabot-navigli-2021-rebel-relation} is a dataset created from Wikipedia abstracts. It consists of an alignment between sentences, Wikipedia hyperlinks and their corresponding Wikidata entities, and relations. REBEL proposed an alignment expanding on~\citet{elsahar-etal-2018-rex},
a pipeline of mention detection, coreference resolution, entity disambiguation and then mapping triplets to each sentence.
\citet{huguet-cabot-navigli-2021-rebel-relation} further filtered false positives using an Natural Language Inference (NLI) model to check if the relation was truly entailed by the text. In this setting, we consider the full $\approx$6M long entity and 857 long relation catalog. We use this dataset for both training and testing. 
Additionally, we employ REBEL to analyze the performance as a function of the number of relations, by simulating different environments pertaining to subsets of the top-$n$ most frequent relations.

\paragraph{Wiki-NRE}\citep{trisedya-etal-2019-neural} is a dataset created from Wikipedia. 
Authors aligned hyperlinks to Wikidata entities as in REBEL but they applied a different filtering: they (i) extracted sentences that contain implicit entity names
using co-reference resolution~\citep{clark-manning-2016-deep}, and (ii) they filtered and assigned relations to sentences using paraphrase detection from different sources~\citep{nakashole-etal-2012-patty,ganitkevitch-etal-2013-ppdb,grycner-weikum-2016-poly}. We used this dataset for both training and testing.

\paragraph{Geo-NRE}\citep{trisedya-etal-2019-neural} is constructed in the same way as Wiki-NRE but from a collection of user reviews on 100 popular landmarks in Australia, instead of Wikipedia. Due to its small size and to compare with the literature, we used this dataset only for testing.

\paragraph{FewRel}\citep{han-etal-2018-fewrel} is also extracted from Wikipedia where Wikidata is the KB. Contrary to the other datasets, FewRel does not provide distant supervision but it is fully annotated by humans. The dataset was first automatically constructed and then filtered as annotators were asked to judge whether the relations are explicitly expressed in the sentences. Each input in FewRel is associated with a single triplet only, and not all of the triplets entailed by it. Therefore, this dataset can be used for precisely measuring recall (but not precision or F1). We employ it only for testing. To simulate a more realistic scenario, we train the models on many relations, and leverage the high quality FewRel data to calculate the performance metrics for the subset of relations annotated.

\section{Performance Metrics} \label{sec:perf_metrics}
We measure standard precision, recall and F1 for all settings.
A fact is regarded as correct if the relation and the two corresponding entities are all correct. More precisely, we denote the set of all predicted triplets of a document $d \in \mathcal{D}$ as $P_d$, and the set of gold triplets as $G_d$. Then:
\begin{align}
    \text{micro-precision} &= \sum_{d \in \mathcal{D}} |P_d \cap G_d| \bigg/ \sum_{d \in \mathcal{D}} |P_d|\;,
\end{align}
and
\begin{align}
    \text{micro-recall} &= \sum_{d \in \mathcal{D}} |P_d \cap G_d| \bigg/ \sum_{d \in \mathcal{D}} |G_d|\;.
\end{align}
Micro scores are useful for measuring the overall performance of a model but they are less informative for imbalanced datasets (\eg, when some entities or relations are disproportionately more present in both training and test sets). Indeed, micro scores assign equal weight to every sample while macro scores assign equal weight to every class. Thus, we also measure macro scores by aggregating per relation type. If we denote $P_d^{(r)}$ and $G_d^{(r)}$ as the predicted and gold set only containing the relation $r \in \mathcal{R}$ of a document $d$, then macro-precision is defined as: 
\begin{align}
    \frac{1}{\mathcal{R}} \sum_{r \in \mathcal{R}}  \left(\; \sum_{d \in \mathcal{D}} |P_d^{(r)} \cap G_d^{(r)}| \bigg/ \sum_{d \in \mathcal{D}} |P_d^{(r)}| \right) \;,
\end{align}
and macro-recall as: 
\begin{align}
   \frac{1}{\mathcal{R}} \sum_{r \in \mathcal{R}}  \left(\; \sum_{d \in \mathcal{D}} |P_d^{(r)} \cap G_d^{(r)}| \bigg/ \sum_{d \in \mathcal{D}} |G_d^{(r)}| \right) \;.
\end{align}

\section{Note on End-to-End Baselines} \label{app:baselines}
We invested a considerable amount of time trying to use a strong end-to-end baseline to compare GenIE with. Unfortunately, most works do not have available or directly usable code. In particular, we first concentrated on SetGenNet~\citep{sui-etal-2021-set} as, to the best of our knowledge, it is the strongest model on the task of cIE. However, the authors do not report a link to the code\footnote{As of October 2021.} in neither the arXiv nor the ACL Antology version of the paper.
We could not find any related repository on GitHub either. For these reasons we were unable to use their method as a baseline for REBEL.

We then focused on the work most similar to SetGenNet, that is the system proposed by~\citet{trisedya-etal-2019-neural}. They released code and we were able to run it. However, the code was incomplete: they included code for training only a part of their system. They start with pre-trained word, entity and relation embeddings, but did not release code for pre-training them. The closest solution we found was using Wikipedia2Vec~\citep{yamada-etal-2020-wikipedia2vec}, which does not include relation embeddings. Besides, the pre-trained word embeddings on the official Wikipedia2Vec website\footnote{\url{https://wikipedia2vec.github.io/wikipedia2vec/pretrained}} do not match the dimensionality used by~\citet{trisedya-etal-2019-neural}. Finally, the authors did not include code to train the ``triple classifier'' of their model. The classifier is instead directly loaded in their code. For these reasons we were unable to use their method as a baseline for REBEL.

\section{Implementation Details} \label{app:implementation_details}
\xhdr{Data}
The train, test and validation splits are either inherited from the original dataset (see Appendix~\ref{app:datasets} for details) or sampled at random. To facilitate reproducibility, we release the exact splits employed in our experiments.

Additionally, we release the curated entity and relation catalogs for both the large and the small schema, in which the redirects have been resolved and each of the QID/PID is paired with a unique, semantically meaningful textual identifier. We hope that this will allow for a fair comparison of future work in which the same evaluation setup can be maintained.

\subsection{GenIE}
\xhdr{Infrastructure}
For training we used a single machine with 24 Intel(R) Xeon(R) CPU E5-2690 v4 @ 2.60GHz processor cores and 441 GB of RAM, equipped with 4 Tesla V100-PCIE-16GB GPUs.

\xhdr{Training}
The models were trained using the Adam optimizer~\citep{kingma2015adam} with a learning rate of 3e-5, 0.1 gradient clipping and a varying weight decay (cf. \Tabref{tab:hyperparameters}). The learning rate is updated using a polynomial decay schedule with an end value of 0. While most of the parameters were left at their default values for BART, the rest were tuned on the respective datasets' validation set, and their corresponding optimal values are given in \Tabref{tab:hyperparameters}.

\begin{table*}
\centering
\resizebox{\textwidth}{!}{
\setlength{\tabcolsep}{5pt}
\begin{tabular}{lrrrccr}
\toprule
{} &  Max steps &  Warm-up steps &  Batch size &  Dropout &  Weight decay & Training time \\
\midrule
GenIE (W)     &      60,000 &           1,000 &          32 &      0.1 &          0.01 & 0.5 GPU days  \\
GenIE (R)     &     100,000 &           5,000 &         384 &      0.1 &          0.01 & 18.5 GPU days \\
GenIE (R + W) &     100,000 &           5,000 &         384 &      0.1 &          0.01 & 20.5 GPU days\\
GenIE - Genre &      50,000 &           3,000 &        2,048 &      0.3 &          0.50 & 11 GPU days \\
GenIE - PLM   &      50,000 &           3,000 &        2,048 &      0.3 &          0.50 & 17 GPU days \\
\midrule
SoTA Rel-class (W)  &      20,000 &           500 &        128 &      0.1 &          0.01 & 0.2 GPU days \\
SoTA Rel-class (R)  &      250,000 &           500 &        128 &      0.1 &          0.01 & 2.5 GPU days \\
SoTA Rel-class (R + W)  &      250,000 &           500 &        128 &      0.1 &          0.01 & 2.5 GPU days \\
SoTA Tri-class (R)  &      50,000 &           500 &        128 &      0.1 &          0.01 & 0.3 GPU days \\
SoTA Tri-class (W)  &      5,000 &           500 &        128 &      0.1 &          0.01 & 0.1 GPU days \\
SoTA Tri-class (R + W)  &      50,000 &           500 &        128 &      0.1 &          0.01 & 0.3 GPU days \\
\bottomrule
\end{tabular}
}
\caption{\textbf{Hyperparameters for the different models.}}
\label{tab:hyperparameters}
\end{table*}

\xhdr{Inference} At test time, we use Constrained Beam Search with 10 beams. We restrict the input and the output sequence to be at most 256 tokens, cutting from the right side if the input is too long. We normalize the log-probabilities by sequence length, and allow for any number of n-gram repetition. The other parameters are kept to their default values for inference with BART.

\subsection{SotA Pipeline}
We described our SotA pipeline system baseline in Sec.~\ref{sec:baselines}. We release code to both train and run inference with the proposed pipeline.
The named entity recognition and the entity disambiguation components were not trained. 
The relation classification module is a linear layer on top of RoBERTa~\citep{liu2019roberta}. We trained it learning rate 3e-4 using the Adam optimizer~\citep{kingma2015adam}. We trained for a maximum number of steps using early stopping on the validation sets. We restrict the input sequence to be at most 128 tokens cutting from the right side if the input is too long. All other hyperparameters are reported in Table~\ref{tab:hyperparameters}.
The triple classification module is also a linear layer on top of RoBERTa~\citep{liu2019roberta} with the same hyperparameters of the relation classification module but we trained for less steps.

\section{Additional Experiments}\label{app:additional_experiments}

\subsection{Analysis of Performance as a Function of the Number of Relations}
\label{app_ssec:eval_n_rel}

\begin{figure}
    \centering
    \includegraphics[width=\linewidth]{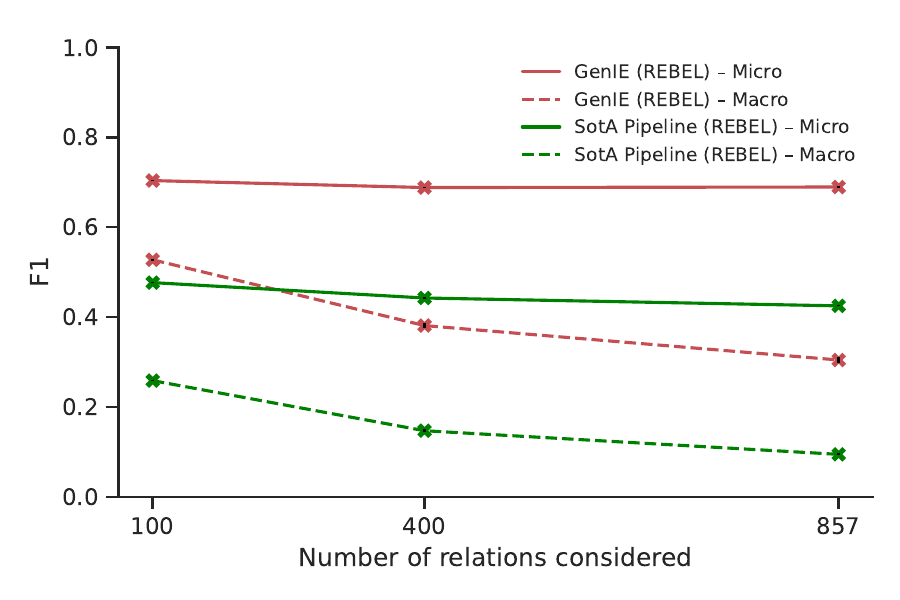}
    \vspace{-22pt}
    \caption{\textbf{Impact of the number of relations in the schema on REBEL.} Micro and macro F1 of both GenIE and the pipeline of SotA components for 3 schema sizes: 100, 400, and 857 relations. The schema is constrained at both training and testing time. Full results (\ie, precision and recall) are reported in \Tabref{tab:appendix_relation-schema} in Appendix~\ref{app:additional_results}.}
    \label{fig:rel_num_exp}
\end{figure}

Previous works focus on small schemas meaning that few relations were considered. Indeed, classification problems on a large set of possible classes become particularly difficult under large class imbalances, which is the case here as shown by \Figref{fig:bucket_plot}.
However, scaling up to larger schemas with more relations is crucial for the models to be useful in downstream tasks.
To measure the scaling ability of GenIE, we create different setups with variable numbers of relations. To create such setups, we start with the REBEL dataset and schema (857 relations) and choose subsets of relations with their associated training data.
In \Figref{fig:rel_num_exp}, we report GenIE and the pipeline baseline F1 for schemas with 100, 400, and 857 relations. To choose a subset of $n$ relations, we take the $n$ most frequent relations to mimic the strategies used by previous works to reduce the schemas~\citep{sui-etal-2021-set}.

We first observe that GenIE is always largely better than the baseline.
The baseline suffers from the same difficulty as previous works; classifying among a large set of relations is hard with large imbalances.
GenIE and the baseline have similar absolute decrease in performance when the number of relations increases, corresponding to a more considerable relative decrease for the baseline. More concretely, GenIE's micro F1-score goes from 70.36 \% for the top 100 relations, to 68.82 \% and 68.93 \% for the top 400 and 857 relation setups, respectively. This translates to a relative decrease of 2 \% only in the first step. For the baseline, the absolute score of 47.67 \% first falls to 44.25 \% and subsequently to 42.5 \% as the number of relations grows. This in turn, is an overall relative drop of almost 11 \%.

Notably, when looking at precision and recall separately (\cf \Tabref{tab:appendix_relation-schema} in Appendix~\ref{app:additional_results}),
GenIE has a slight proportional decrease of 1-2 absolute points, both in precision and recall, which reflects the increased difficulty of the task due to larger number of relations. The baseline exhibits a similar drop in precision, but a much more significant drop in the recall of almost 10 absolute or 16 point relative. This suggests that the baseline simply ignores most of the relations with lower occurrence counts, which is consistent with the results in \Secref{fig:bucket_plot}, and the hypothesis that the relation classification task is a bottleneck for effectively scaling the baseline system to a large number of relations.

We already have to deploy several techniques to help the baseline better deal with these issues (see \Secref{sec:baselines}), while GenIE, thanks to its generative autoregressive formulation, can effectively scale and manage the inherent imbalances of the task much more naturally.

\clearpage
\onecolumn
\section{Additional Results}\label{app:additional_results}

\begin{table*}[h]
    \centering
    \begin{tabular}{lcccccc}
        \toprule
        & \multicolumn{3}{c}{\textbf{Wiki-NRE}} & \multicolumn{3}{c}{\textbf{Geo-NRE}} \\
        & Precision & Recall & F1 & Precision & Recall & F1 \\
        \midrule
        \midrule
        \textbf{\textit{Pipeline baselines}} \\
        \hspace{4mm} AIDA + MinIE  & 36.72 & 48.56 & 41.82 & 35.74 & 39.01 & 37.30 \\
        \hspace{4mm} NeuralEL + MinIE  & 35.11 & 39.67 & 37.25 & 36.44 & 38.11 & 37.26 \\
        \hspace{4mm} AIDA + ClauseIE  & 36.17 & 47.28 & 40.99 & 35.31 & 39.51 & 37.29 \\
        \hspace{4mm} NerualEL + ClauseIE  & 34.45 & 37.86 & 36.07 & 35.63 & 37.91 & 36.73 \\
        \hspace{4mm} AIDA + CNN  & 40.35 & 35.03 & 37.50 & 37.15 & 31.65 & 34.18 \\
        \hspace{4mm} NeuralEL + CNN  & 36.89 & 35.21 & 36.03 & 37.81 & 30.05 & 33.49 \\
        \midrule
        \textbf{\textit{Encoder-decoder baselines}} \\
        \hspace{4mm} Single Attention & 45.91 & 38.36 & 41.80 & 40.10 & 39.12 & 39.60 \\
        \hspace{4mm} Single Attention (+pre-trained)  & 47.25 & 40.53 & 43.63 & 43.14 & 43.11 & 43.12 \\
        \hspace{4mm} Single Attention (+beam)  & 60.56 & \underline{52.31} & 56.13 & 58.69 & 48.51 & 53.12 \\
        \hspace{4mm} Single Attention (+triplet classifier)  & \textbf{73.78} & 50.13 & \underline{59.70} & \underline{67.04} & 53.01 & 59.21 \\
        \hspace{4mm} Transformer  & 46.28 & 38.97 & 42.31 & 45.75 & 46.20 & 45.97 \\
        \hspace{4mm} Transformer (+pre-trained)  & 47.48 & 40.91 & 43.95 & 48.41 & 48.31 & 48.36 \\
        \hspace{4mm} Transformer (+beam)  & 58.29 & 50.25 & 53.97 & 61.81 & \underline{61.61} & 61.71 \\
        \hspace{4mm} Transformer (+triplet classifier)  & \underline{73.07} & 48.66 & 58.42 & \textbf{71.24} & 57.61 & \underline{63.70} \\
        \midrule
        \textbf{Our pipeline baseline} & 67.43 & \textbf{54.22} & \textbf{60.11} &  64.60 & \textbf{64.05} & \textbf{64.32} \\
        \bottomrule
    \end{tabular}
    \caption{
    \textbf{Baselines comparison.} All results are taken from from~\citet{trisedya-etal-2019-neural}. Encoder-decoder baseline are proposed by the authors and other pipeline baseline include an NER and an ED system AIDA~\citep{hoffart-etal-2011-robust} or NeuralEL~\citep{kolitsas-etal-2018-end} and then a relation extraction system CNN~\citep{lin-etal-2016-neural}, MiniE~\citep{gashteovski-etal-2017-minie}, or ClausIE~\citep{del2013clausie}. 
    Best results are highlighted in \textbf{bold} and second best are \underline{underlined}. Our pipeline baseline scores the best or on pair among these other methods.
    }
    \label{tab:baseline_results}
\end{table*}

\begin{figure*}
    \centering
    \includegraphics[width=\linewidth]{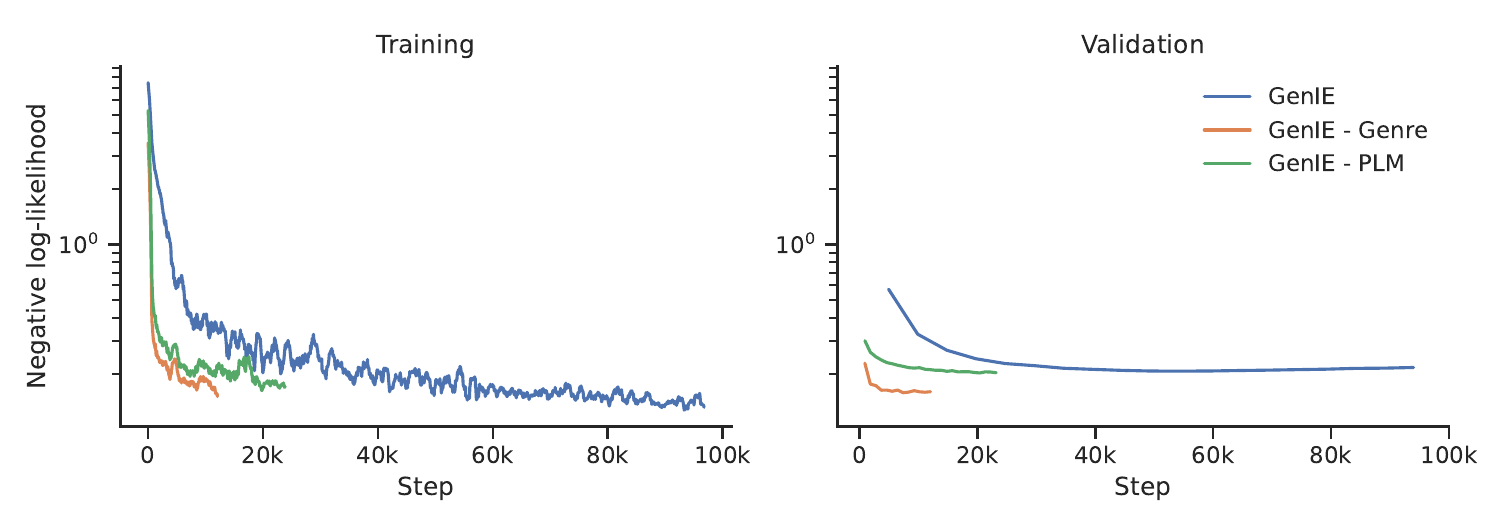}
    \vspace{-22pt}
    \caption{\textbf{Training and validation loss curves for different initialization of our model.} GenIE starts from a random initialization, GenIE – PLM fine-tunes a BART pre-trained language model, while GenIE - GENRE is initialized with a pre-trained autoregressive entity linking model by~\citet{cao2021autoregressive}.}
    \label{fig:training_loss}
\end{figure*}

\begin{table*}[h]
\centering
\resizebox{\textwidth}{!}{
\setlength{\tabcolsep}{5pt}
\begin{tabular}{@{}lccc|ccc|ccc@{}}
\toprule
& \multicolumn{3}{c}{\textbf{REBEL (top 100 Relations)}} & \multicolumn{3}{c}{\textbf{REBEL (top 400 Relations)}} & \multicolumn{3}{c}{\textbf{REBEL (857 Relations)}} \\
& Precision & Recall & F1 & Precision & Recall & F1 & Precision & Recall & F1 \\
\midrule
\midrule
\emph{\textbf{Micro}} \\ 
\hspace{4mm} GenIE & 68.76 {\scriptsize± 0.12} & 72.05 {\scriptsize± 0.13} & 70.36 {\scriptsize± 0.10} & 67.10 {\scriptsize± 0.13} & 70.62 {\scriptsize± 0.15} & 68.82 {\scriptsize± 0.12} & 68.02 {\scriptsize± 0.15} & 69.87 {\scriptsize± 0.14} & 68.93 {\scriptsize± 0.12} \\
\hspace{4mm} SotA Pipeline & 44.76 {\scriptsize± 0.17} & 50.99 {\scriptsize± 0.17} & 47.67 {\scriptsize± 0.16} & 38.98 {\scriptsize± 0.13} & 51.18 {\scriptsize± 0.12} & 44.25 {\scriptsize± 0.11} & 43.30 {\scriptsize± 0.15} & 41.73 {\scriptsize± 0.13} & 42.50 {\scriptsize± 0.13} \\
\midrule
\emph{\textbf{Micro}} \\ 
\hspace{4mm} GenIE & 52.26 {\scriptsize± 0.25} & 54.13 {\scriptsize± 0.27} & 52.75 {\scriptsize± 0.24} & 41.50 {\scriptsize± 0.66} & 38.53 {\scriptsize± 0.57} & 38.12 {\scriptsize± 0.51} & 33.90 {\scriptsize± 0.73} & 30.48 {\scriptsize± 0.65} & 30.46 {\scriptsize± 0.62} \\
\hspace{4mm} SotA Pipeline & 27.41 {\scriptsize± 0.27} & 31.05 {\scriptsize± 0.18} & 25.87 {\scriptsize± 0.15} & 16.94 {\scriptsize± 0.63} & 19.00 {\scriptsize± 0.36} & 14.73 {\scriptsize± 0.37} & 12.20 {\scriptsize± 0.35} & 10.44 {\scriptsize± 0.22} & 9.48 {\scriptsize± 0.21} \\
\bottomrule                            
\end{tabular}
}
\caption{\textbf{Impact of the number of relations in the schema on REBEL.} The schema is constrained at both training and testing time.}
\label{tab:appendix_relation-schema}
\end{table*}



\end{document}